\documentclass[runningheads]{llncs}

\usepackage{eccv}

\usepackage{eccvabbrv}

\usepackage{graphicx}
\usepackage{booktabs}
\usepackage{float}
\usepackage{tikz}
\usetikzlibrary{arrows.meta,backgrounds,calc,fit,positioning}

\usepackage[accsupp]{axessibility}  

\usepackage[pagebackref,breaklinks,colorlinks,citecolor=eccvblue]{hyperref}
\usepackage{hyperref}

\usepackage{orcidlink}

\begin{document}

\title{UniTraffic-Agent: Unified Traffic Video Reasoning for AI City Challenge 2026 Track 3 with Two Out-of-Domain Evaluations}

\titlerunning{UniTraffic-Agent}

\author{
Peng Li\inst{1} \and
Qianqian Xu\inst{2,3}$^\star$ \and
Shilong Bao\inst{2} \and
Yangbangyan Jiang\inst{4} \and
Qingming Huang\inst{1,2}\thanks{Corresponding authors.}
}

\authorrunning{P.~Li et al.}

\institute{
School of Computer Science and Technology, University of Chinese Academy of Sciences (UCAS), Beijing, China
\and
State Key Laboratory of AI Safety, Institute of Computing Technology (ICT), Chinese Academy of Sciences (CAS), Beijing, China
\and
Beijing Academy of Artificial Intelligence (BAAI), Beijing, China
\and
School of Artificial Intelligence and Robotics, Hunan University, Changsha, China
\\
\email{lipeng251@mails.ucas.ac.cn, qmhuang@ucas.ac.cn, xuqianqian@ict.ac.cn, baoshilong@ict.ac.cn, jiangyangbangyan@hnu.edu.cn}
}

\maketitle

\begin{abstract}
Traffic video understanding has become an important problem in intelligent transportation, as road videos provide direct evidence for accidents, violations, and interactions between vehicles and vulnerable road users. A useful system should explain how a traffic event develops, why it happens, and when the relevant interaction occurs, yet this remains difficult for multimodal large language models (MLLMs) because traffic videos contain sparse events and varied viewpoints. We introduce UniTraffic-Agent, the MR-CAS solution for Track~3 of the 10th AI City Challenge, which includes Traffic Anomaly Reasoning (TAR) and two out-of-domain evaluations: FETV for fisheye traffic events and PSI-VQA for pedestrian intention reasoning. UniTraffic-Agent follows an observe--reason--act--verify workflow that samples timestamped visual evidence, reasons over all questions from the same clip in one request, and converts responses through task-specific action adapters. On the official Public leaderboards, MR-CAS ranks 16th on TAR with a score of 0.5780, 2nd on FETV with 0.4884, and 4th on PSI-VQA with 64.4161. The code is available at \url{https://github.com/Roclp/UniTraffic-Agent}.
  \keywords{Traffic video understanding \and Multimodal large language models \and Agentic reasoning \and Out-of-domain generalization}
\end{abstract}

\section{Introduction}
\label{sec:intro}
Traffic video understanding has attracted increasing attention with the widespread deployment of cameras on urban roads, at intersections, along highways, and in vehicles. Recent traffic--language systems, including TrafficVLM~\cite{trafficvlm2024}, TrafficVILA~\cite{trafficvila2025}, and STER-VLM~\cite{stervlm2025}, reflect the growing interest in interpreting safety-critical events from video. Such events may involve vehicles running red lights, pedestrians hesitating at curbs, or drivers yielding to other road users. Understanding them requires a system not only to recognize relevant actors and road conditions, but also to track their behavior over time and relate observed motion to the eventual outcome~\cite{jaad2017,pie2019,vadr1}.

MLLMs provide a flexible interface for this problem by jointly processing sampled frames and natural-language instructions~\cite{gpt4v2023,gemini2023,llava2023,videochatgpt2024}. However, traffic videos pose challenges beyond conventional image and short-video question answering. Critical evidence may occur in only a few frames, viewpoints vary considerably across surveillance cameras, fisheye lenses, and dashcams, and a single clip may be associated with multiple questions. Moreover, answering each question independently can lead to inconsistent predictions~\cite{qvhighlights2021,timechat2024,domainaware2025,yang2023revisiting,bao2025towards}.

Track~3 of the 10th AI City Challenge~\cite{Tang26AICity26} provides a benchmark for traffic video understanding. Its main Traffic Anomaly Reasoning (TAR) task covers multiple question types. Two optional out-of-domain tasks further evaluate generalization across cameras. Fisheye Traffic Event Understanding (FETV)~\cite{fetv2026} requires structured violation records containing actor, trajectory, road, environmental, and temporal information. Pedestrian Scenario Intention Visual Question Answering (PSI-VQA) evaluates crossing-intent prediction and the localization of supporting evidence for a designated pedestrian in dashcam videos~\cite{psi2026,bao2024improved}. 

To address this challenge, we introduce \textbf{UniTraffic-Agent}, a unified traffic-video agent based on an \emph{observe--reason--act--verify} workflow. During observation, the agent constructs a compact frame set that combines global video coverage with samples near question-specific timestamps. During reasoning, all questions and output fields associated with a clip are processed jointly to establish a shared event interpretation. Task-specific adapters then convert this interpretation into the official format required by each task. Finally, a verifier checks identifiers and retries unresolved cases using cached visual evidence.

Our contributions are summarized as follows:
\begin{itemize}
    \item \textbf{Unified traffic-video agent framework.}
    We introduce UniTraffic-Agent, which supports heterogeneous traffic-understanding tasks across surveillance, fisheye, and dashcam videos through an observe--reason--act--verify workflow.

    \item \textbf{Timestamp-aware observation and reasoning.}
    We combine global frame coverage with question-specific temporal evidence and perform clip-level joint reasoning to improve prediction consistency.

    \item \textbf{Task-specific action adapters.}
    We develop task-specific action adapters and validation procedures for different tasks. MR-CAS ranks 2nd on FETV and 4th on PSI-VQA on the official Public leaderboards.
\end{itemize}

\section{Related Work}
\label{sec:related}
\paragraph{Multimodal large language models and agents.}
In recent years, multimodal large language models (MLLMs) have shown strong capabilities in image question answering and video description, including GPT-4V~\cite{gpt4v2023}, Gemini~\cite{gemini2023}, LLaVA~\cite{llava2023}, and Video-ChatGPT~\cite{videochatgpt2024}. VideoCLIP~\cite{videoclip2021}, VideoCoCa~\cite{videococa2023}, and TimeChat~\cite{timechat2024} further study video--text representation and temporal grounding. Traffic-oriented systems adapt MLLMs to road scenes through phase-aware inputs, high-resolution views, reference examples, specialized prompts, or multi-agent reasoning~\cite{trafficvlm2024,trafficvila2025,stervlm2025,domainaware2025,multiagent2025,hua2025openworldauc,bao2025aucpro}. SpatialAgent~\cite{spatialagent2025} shows how an LLM agent can coordinate perception tools for AI City spatial QA. UniTraffic-Agent follows this agent-based direction for traffic video reasoning across CCTV, fisheye, and dashcam settings.

\paragraph{Traffic anomaly and temporal reasoning.} Traffic anomaly research covers detection~\cite{ucfcrime,tadbenchmark,sotad}, weakly supervised localization~\cite{tadlocalization}, accident understanding~\cite{accidentbench}, and causal reasoning~\cite{vadr1}. Query-conditioned models such as QVHighlights~\cite{qvhighlights2021} and TimeChat~\cite{timechat2024} further connect language queries to temporal video intervals. Track~3 extends these settings by requiring one system to handle multiple question types and camera domains: TAR supports diverse anomaly-reasoning questions for the same clip~\cite{hua2024reconboost}, FETV builds on fisheye traffic perception~\cite{fisheye8k2023,fetv2026}, and PSI-VQA extends pedestrian crossing benchmarks~\cite{jaad2017,pie2019,psi2026} with cue explanations and decision-relevant intervals~\cite{bao2022rethinking}. These tasks motivate our video-level inference and task-specific action adapters.

\section{UniTraffic-Agent}
\label{sec:method}
\begin{figure*}[t]
\centering
    \includegraphics[width=1\linewidth]{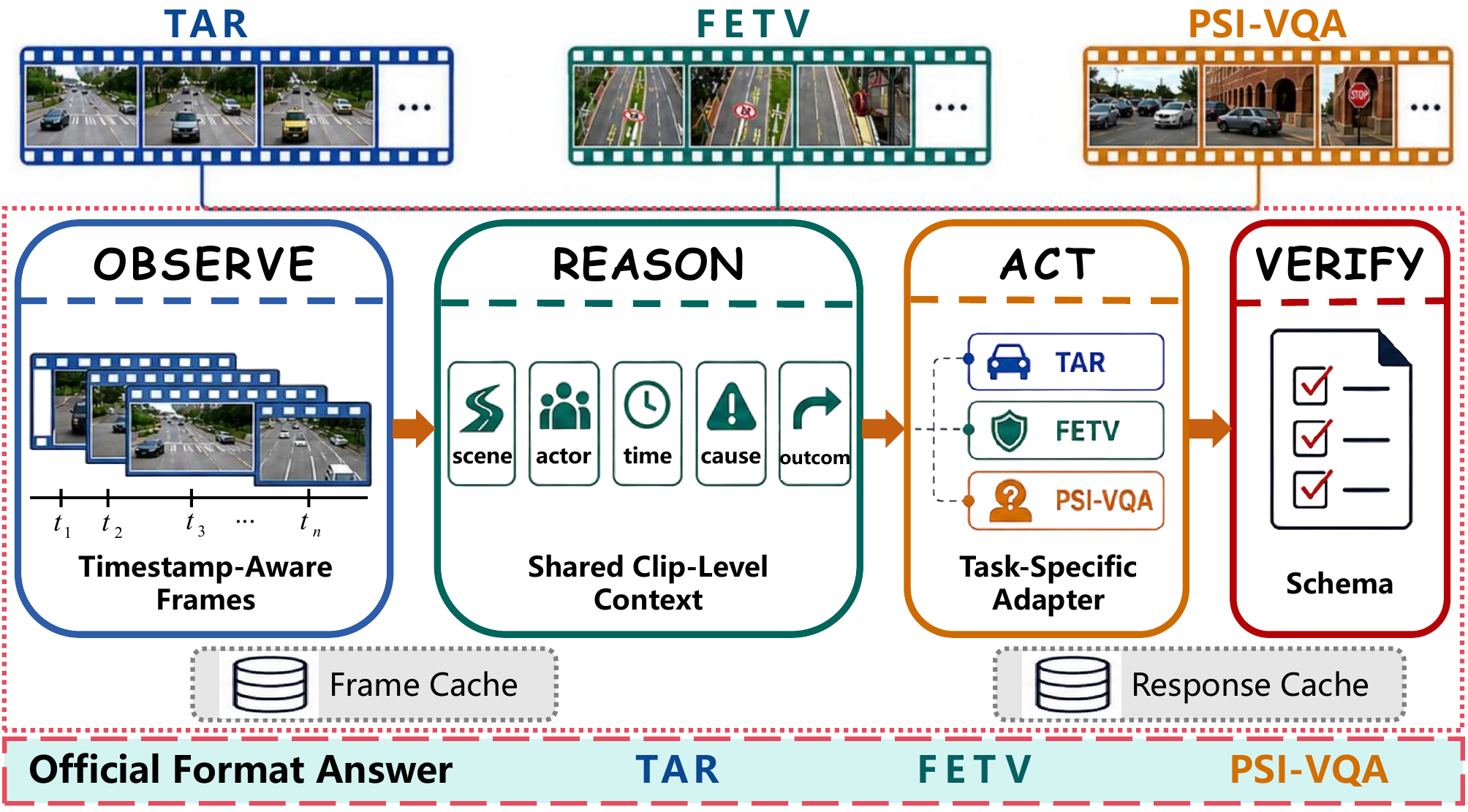}
    \caption{Overview of UniTraffic-Agent. The observe--reason--act--verify workflow shares timestamp-aware observation, clip-level event reasoning, and output verification across TAR, FETV, and PSI-VQA, while task-specific action adapters convert the shared event interpretation into each official submission format.}
\label{fig:pipeline}
\end{figure*}

UniTraffic-Agent is designed as a unified inference workflow for heterogeneous traffic-video tasks. Instead of treating each question or output field as an isolated query, the system uses the video clip as the basic reasoning unit. For each clip, it first builds a timestamped visual observation, then asks an MLLM to infer a shared event context, including the relevant actors, road layout, temporal evolution, and outcome. This shared context is finally mapped to benchmark-specific predictions by task-specific action adapters. As shown in Figure~\ref{fig:pipeline}, TAR, FETV, and PSI-VQA use the same observation strategy, model interface, caching mechanism, and verification procedure, while their adapters specify the required input schema, answer space, formatting examples, and submission format.

\subsection{Timestamp-Aware Observation}
Processing every frame of a video is impractical for a hosted MLLM because of input-length and computational constraints. We therefore construct a compact frame set that balances global coverage with task-specific temporal evidence. The sampler first selects $G$ frames uniformly distributed over the entire clip. It then adds a duration-adaptive temporal grid for broader temporal coverage. For TAR and PSI-VQA, questions may provide explicit timestamps; in such cases, we additionally sample a local neighborhood from $-1$ to $+1$ seconds around each timestamp anchor. If the candidate set exceeds the frame budget $M$, endpoints and frames closest to timestamp anchors are kept first, and the remaining slots are filled from the regular grid. Duplicate frames are removed and the final frames are ordered chronologically. Each frame is paired with its timestamp. The prompt states that adjacent inputs may be separated by several seconds, and frames near timestamp anchors are sent with higher visual detail. Decoded JPEGs are cached so retries and recovery runs use identical visual evidence.


\subsection{Video-Level Event Reasoning}
Traffic clips in Track~3 contain multiple questions and output fields that refer to the same underlying event. UniTraffic-Agent therefore builds a shared event context before producing task-specific answers. For each clip, the model is instructed to examine the road layout, identify the relevant actors, and trace the event from its initial state to its final outcome. It then answers all associated questions or fields in a single request using this shared context. This video-level reasoning reduces inconsistencies in actor identity, causal explanations, and temporal boundaries across outputs that describe the same event.

\subsection{Task-Specific Action Adapters}
Although the three benchmarks share the same observation and reasoning protocol, they differ substantially in their answer spaces and submission formats. We therefore implement a task-specific action adapter for each benchmark to map the shared reasoning output to the required submission format.

\paragraph{TAR.} The TAR adapter jointly handles all question types for a video. It constrains BCQ and MCQ answers while supporting free-form descriptions, causal explanations, summaries, and temporal intervals. The resulting predictions are matched to official identifiers and written to the required CSV format.

\paragraph{FETV.} The FETV adapter generates a structured 13-field violation record covering the violator, trajectory, road context, timestamp, and event description. It distinguishes positions in the $3\times3$ grid from lane indices defined by the direction of travel, normalizes categorical values, and produces the official JSON output.

\paragraph{PSI-VQA.} The PSI-VQA adapter tracks the pedestrian marked by a red box and distinguishes observed motion from inferred crossing intent. It produces BCQ, MCQ, visual-cue, and temporal answers. Decision-relevant intervals cover the period from when the pedestrian begins to affect the driving decision until the relevant action ends or the pedestrian leaves the field of view. Predictions are converted to the official CSV format.

\subsection{Verification and Recovery}
Verification restores official identifiers, normalizes categorical values, validates temporal intervals, and checks output completeness without modifying semantically valid answers. Frame and response caches avoid redundant computation and preserve successful predictions during recovery. Raw responses and validation results are retained for error analysis.

\section{Experiments}
\label{sec:experiments}

\subsection{Datasets}

\paragraph{TAR.} The TAR test set~\cite{Tang26AICity26} contains 960 human-verified questions for 80 CCTV clips trimmed from 17 public YouTube videos. The training data are collected from eight public traffic and anomaly datasets~\cite{vadr1,tadlocalization,accidentbench,tadbenchmark,sotad,ucfcrime}, resulting in substantial variation in viewpoint, duration, and event type.

\paragraph{FETV.} The FETV test set~\cite{fetv2026} contains 200 fisheye clips with traffic violations and normal traffic scenes. Each clip is annotated with the violator, trajectory, road geometry, environment, event time, and event description.

\paragraph{PSI-VQA.} The PSI-VQA test set~\cite{psi2026} contains 328 questions for 40 dashcam clips. All questions concern a marked pedestrian and cover crossing intent, supporting visual cues, multiple-choice reasoning, or temporal localization.

\subsection{Evaluation Metrics}
Following the official protocol~\cite{Tang26AICity26}, we evaluate UniTraffic-Agent on TAR, FETV, and PSI-VQA using the AI City Challenge evaluation system.

\paragraph{TAR.}
For TAR, BCQ and MCQ are evaluated by accuracy, while seven open-ended task types are evaluated by BERTScore F1~\cite{bertscore}. The final TAR score is the unweighted mean over the nine scored task types, where $\mathcal{T}_{\mathrm{open}}$ denotes the seven open-ended task types:
\begin{equation}
S_{\mathrm{TAR}} =
\frac{1}{9}
\left(
S_{\mathrm{BCQ}} +
S_{\mathrm{MCQ}} +
\sum_{t \in \mathcal{T}_{\mathrm{open}}} S_t
\right).
\end{equation}

\paragraph{FETV.}
FETV scores structured violation attributes and event descriptions~\cite{fetv2026}. The structured fields are summarized by the official MacroF1 term, while the description is evaluated by normalized CIDEr and BERTScore. The final FETV score follows the official weighted formula:
\begin{equation}
S_{\mathrm{FETV}} =
0.25 \cdot \mathrm{CIDEr}_{\mathrm{norm}}
+
0.25 \cdot \mathrm{BERTScore}
+
0.5 \cdot \mathrm{MacroF1}.
\end{equation}

\paragraph{PSI-VQA.}
PSI-VQA evaluates four pedestrian-intent subtasks~\cite{psi2026}: Macro-F1 for PSI-T1, cue-level F1 for PSI-T2, accuracy for PSI-T3, and temporal mIoU for PSI-T4. The four normalized subtask scores are combined with equal weight:
\begin{equation}
S_{\mathrm{PSI}} =
0.25 \cdot \mathrm{PSI\text{-}T1}
+
0.25 \cdot \mathrm{PSI\text{-}T2}
+
0.25 \cdot \mathrm{PSI\text{-}T3}
+
0.25 \cdot \mathrm{PSI\text{-}T4}.
\end{equation}

\subsection{Implementation Details}
For all three tasks, each video is processed in a single request with at most 32 sampled frames, including 16 globally distributed frames. Frames are cached as JPEG images with quality 100 and a maximum side length of 768 pixels, and frames near question-provided timestamps use the high-detail setting.Video‑level requests run in parallel for throughput. All API calls use temperature 0 with up to three transport retries. We use \texttt{gpt-5.5} as the primary model. Cases that remain missing or unparsable after API or validation failures are recovered with \texttt{gpt-5.4} using the same prompt and cached frames. Public training annotations are used only to construct answer-format examples. We do not use private data, manually annotate the test set, or edit individual test predictions.

\begin{table}[t]
\centering
\caption{Official Public leaderboard results for MR-CAS and the leading entries on TAR, FETV, and PSI-VQA.}
\label{tab:leaderboard}
\begin{tabular}{lccc}
\toprule
Evaluation & Rank & MR-CAS & Leader \\
\midrule
TAR & 16 & 0.5780 & 0.6788 \\
FETV & 2 & 0.4884 & 0.4891 \\
PSI-VQA & 4 & 64.4161 & 70.6397 \\
\bottomrule
\end{tabular}
\end{table}

\subsection{Official Challenge Results}

Table~\ref{tab:leaderboard} summarizes our official Public leaderboard results together with the leading score for each task. MR-CAS ranks 2nd on FETV, 4th on PSI-VQA, and 16th on TAR, with the FETV result within 0.0007 of 1st place.

\begin{table}[t]
\centering
\caption{TAR final and component results for MR-CAS and the leading Public entry. Gap denotes MR-CAS minus Leader.}
\label{tab:tar-results}
\begin{tabular}{lccc}
\toprule
Metric & MR-CAS & Leader & Gap \\
\midrule
Official mean & 0.5780 & \textbf{0.6788} & -0.1008 \\
BCQ & 0.9187 & \textbf{1.0000} & -0.0813 \\
MCQ & \textbf{0.9500} & \textbf{0.9500} & 0.0000 \\
BCQ-OE & 0.5585 & \textbf{0.6686} & -0.1101 \\
MCQ-OE & 0.9604 & \textbf{0.9693} & -0.0089 \\
Open-QA & 0.4042 & \textbf{0.4986} & -0.0944 \\
Causal linkage & 0.4192 & \textbf{0.5310} & -0.1118 \\
Scene description & 0.2667 & \textbf{0.4373} & -0.1706 \\
Temporal description & 0.3248 & \textbf{0.5137} & -0.1889 \\
Summarization & 0.3992 & \textbf{0.5409} & -0.1417 \\
Temporal localization & 0.6936 & \textbf{0.7803} & -0.0867 \\
\bottomrule
\end{tabular}
\end{table}

\begin{table}[t]
\centering
\caption{FETV final and component results for MR-CAS and the leading Public entry. Gap denotes MR-CAS minus Leader.}
\label{tab:fetv-results}
\begin{tabular}{lccc}
\toprule
Metric & MR-CAS & Leader & Gap \\
\midrule
Final score & 0.4884 & \textbf{0.4891} & -0.0007 \\
Categorical mean & 0.5358 & \textbf{0.5612} & -0.0254 \\
Description & \textbf{0.4411} & 0.4171 & +0.0240 \\
Violation type & 0.2302 & \textbf{0.2542} & -0.0240 \\
Violator type & \textbf{0.4543} & 0.4442 & +0.0101 \\
Color & 0.1911 & \textbf{0.1981} & -0.0070 \\
Initial position & \textbf{0.2647} & 0.1942 & +0.0705 \\
Final position & \textbf{0.2734} & 0.2123 & +0.0611 \\
Initial lane & 0.1948 & \textbf{0.2218} & -0.0270 \\
Final lane & \textbf{0.2507} & 0.2149 & +0.0358 \\
Intersection type & 0.5749 & \textbf{1.0000} & -0.4251 \\
Weather & \textbf{1.0000} & \textbf{1.0000} & 0.0000 \\
Light condition & \textbf{1.0000} & \textbf{1.0000} & 0.0000 \\
Date & \textbf{1.0000} & \textbf{1.0000} & 0.0000 \\
Time & \textbf{0.9950} & \textbf{0.9950} & 0.0000 \\
\bottomrule
\end{tabular}
\end{table}

\subsubsection{TAR}

Table~\ref{tab:tar-results} shows that MR-CAS performs well on constrained questions, matching the leader on MCQ accuracy and reaching 0.9604 on MCQ-OE F1. The main gap comes from long-form tasks, including scene description, temporal description, and summarization. These tasks require not only identifying the anomalous event, but also selecting reference-aligned details and temporal abstractions. The relatively strong temporal mIoU suggests that the model often localizes the event, but still struggles to express the evidence at the same level of granularity as the reference answer.

\subsubsection{FETV}

MR-CAS achieves 0.4884 on FETV, only 0.0007 below the leading Public score. Table~\ref{tab:fetv-results} shows that the system is competitive on description, violator type, image-position fields, and final lane, indicating that timestamp-aware observation and the FETV action adapter help locate and track the violating actor. The main weakness is road-topology reasoning, especially intersection type, where fisheye distortion makes geometric interpretation difficult.

\subsubsection{PSI-VQA}

Table~\ref{tab:psi-results} reports the PSI-VQA component results. MR-CAS ranks 4th overall and scores higher than the leading entry on Open-QA Cue-F1, indicating that the model capture useful visual evidence for the target pedestrian. However, BCQ and temporal localization scores remain lower than the leader. This suggests that cue recognition alone is not sufficient for PSI-VQA, where accurate performance also depends on reliable crossing-intent judgment and alignment between the predicted interval and the driver-relevant decision window.

\begin{table}[t]
\centering
\caption{PSI-VQA final and component results for MR-CAS and the leading Public entry. Gap denotes MR-CAS minus Leader.}
\label{tab:psi-results}
\begin{tabular}{lccc}
\toprule
Metric & MR-CAS & Leader & Gap \\
\midrule
Final score & 64.4161 & \textbf{70.6397} & -6.2236 \\
BCQ & 0.5934 & \textbf{0.7084} & -0.1150 \\
BCQ acc. & 0.6000 & \textbf{0.7273} & -0.1273 \\
Open-QA Cue & \textbf{0.6389} & 0.5833 & +0.0556 \\
MCQ & 0.7692 & \textbf{0.7912} & -0.0220 \\
Temporal localization & 0.5751 & \textbf{0.7427} & -0.1676 \\
\bottomrule
\end{tabular}
\end{table}

\subsubsection{Cross-Domain Analysis}

FETV and PSI-VQA provide two out-of-domain evaluations beyond TAR: fisheye intersection videos with structured violation records, and dashcam pedestrian-intent videos with a marked target actor. Without task-specific fine-tuning, UniTraffic-Agent ranks 2nd on FETV and 4th on PSI-VQA, showing the effectiveness of transfer to fisheye and dashcam settings.

\section{Conclusion}
\label{sec:conclusion}

We introduce UniTraffic-Agent for Track~3 of the 10th AI City Challenge. By combining timestamp-aware observation, clip-level event reasoning, and task-specific action adapters, UniTraffic-Agent shows strong out-of-domain generalization, with MR-CAS ranking 2nd on FETV and 4th on PSI-VQA, while also performing well on selected TAR and PSI-VQA components. Remaining errors mainly arise from reference-aligned long-form generation, fisheye road geometry, pedestrian intent prediction, and temporal boundary estimation. These results suggest that explicit actor tracking and geometry-aware temporal reasoning are promising directions for future traffic-video agents.

\bibliographystyle{splncs04}
\bibliography{main}

@InProceedings{Tang26AICity26,
  author    = {Tang, Zheng and Wang, Shuo and Anastasiu, David C. and Chang, Ming-Ching and others},
  title     = {The 10th {AI City Challenge}},
  booktitle = {ECCV Workshops},
  year      = {2026},
  address   = {Malm{"o}, Sweden}
}

@misc{gpt4v2023,
  author       = {{OpenAI}},
  title        = {{GPT-4V(ision) System Card}},
  year         = {2023},
  howpublished = {\url{https://openai.com/index/gpt-4v-system-card/}}
}

@article{gemini2023,
  title={Gemini: a family of highly capable multimodal models},
  author={Team, Gemini and Anil, Rohan and Borgeaud, Sebastian and Alayrac, Jean-Baptiste and Yu, Jiahui and Soricut, Radu and Schalkwyk, Johan and Dai, Andrew M and Hauth, Anja and Millican, Katie and others},
  journal={arXiv preprint arXiv:2312.11805},
  year={2023}
}

@inproceedings{llava2023,
  author    = {Liu, Haotian and Li, Chunyuan and Wu, Qingyang and Lee, Yong Jae},
  title     = {Visual Instruction Tuning},
  booktitle = {Advances in Neural Information Processing Systems},
  year      = {2023}
}

@inproceedings{videochatgpt2024,
  author    = {Maaz, Muhammad and Rasheed, Hanoona and Khan, Salman and Khan, Fahad Shahbaz},
  title     = {{Video-ChatGPT}: Towards Detailed Video Understanding via Large Vision and Language Models},
  booktitle = {Proceedings of the 62nd Annual Meeting of the Association for Computational Linguistics (Volume 1: Long Papers)},
  pages     = {12585--12602},
  year      = {2024}
}

@inproceedings{videoclip2021,
  author    = {Xu, Hu and Ghosh, Gargi and Huang, Po-Yao and Okhonko, Dmytro and Aghajanyan, Armen and Metze, Florian and Zettlemoyer, Luke and Feichtenhofer, Christoph},
  title     = {{VideoCLIP}: Contrastive Pre-training for Zero-shot Video-Text Understanding},
  booktitle = {Proceedings of the Conference on Empirical Methods in Natural Language Processing},
  pages     = {6787--6800},
  year      = {2021}
}

@article{videococa2023,
  author  = {Yan, Shen and Zhu, Tao and Wang, Zirui and Cao, Yuan and Zhang, Mi and Ghosh, Soham and Wu, Yonghui and Yu, Jiahui},
  title   = {{VideoCoCa}: Video-Text Modeling with Zero-Shot Transfer from Contrastive Captioners},
  journal = {arXiv preprint arXiv:2212.04979},
  year    = {2022}
}

@inproceedings{timechat2024,
  title={Timechat: A time-sensitive multimodal large language model for long video understanding},
  author={Ren, Shuhuai and Yao, Linli and Li, Shicheng and Sun, Xu and Hou, Lu},
  booktitle={2024 IEEE/CVF Conference on Computer Vision and Pattern Recognition (CVPR)},
  pages={14313--14323},
  year={2024},
  organization={IEEE}
}

@article{qvhighlights2021,
  title={Detecting moments and highlights in videos via natural language queries},
  author={Lei, Jie and Berg, Tamara L and Bansal, Mohit},
  journal={Advances in Neural Information Processing Systems},
  volume={34},
  pages={11846--11858},
  year={2021}
}

@inproceedings{trafficvlm2024,
  author    = {Dinh, Quang Minh and Ho, Minh Khoi and Dang, Anh Quan and Tran, Hung Phong},
  title     = {{TrafficVLM}: A Controllable Visual Language Model for Traffic Video Captioning},
  booktitle = {Proceedings of the IEEE/CVF Conference on Computer Vision and Pattern Recognition Workshops},
  pages     = {7134--7143},
  year      = {2024}
}

@inproceedings{trafficvila2025,
  author    = {Park, Beomseok and Yang, Wanzhao and Yuan, Sifan and Anwar, Syed Muhammad and Marsic, Ivan},
  title     = {{TrafficVILA}: A Multimodal Framework for Traffic Safety Description and Analysis},
  booktitle = {Proceedings of the IEEE/CVF International Conference on Computer Vision Workshops},
  pages     = {5460--5468},
  year      = {2025}
}

@inproceedings{stervlm2025,
  title={STER-VLM: Spatio-Temporal With Enhanced Reference Vision-Language Models},
  author={Nguyen-Nhu, Tinh-Anh and Minh, Triet Dao Hoang and To-Thanh, Dat and Le-Gia, Phuc and Vo-Lan, Tuan and Nguyen, Tien-Huy},
  booktitle={2025 IEEE/CVF International Conference on Computer Vision Workshops (ICCVW)},
  pages={5516--5525},
  year={2025},
  organization={IEEE}
}

@inproceedings{domainaware2025,
  title={Domain-aware enhancements to vision-language models for urban traffic safety question answering},
  author={Ha, Vu Thanh Dat and Tran, Tuan Huy and Dong, Gang Thep and Chu, Ngoc Chien and Vu, Huan and Nguyen, Tien Cuong},
  booktitle={2025 IEEE/CVF International Conference on Computer Vision Workshops (ICCVW)},
  pages={5425--5433},
  year={2025},
  organization={IEEE}
}

@inproceedings{multiagent2025,
  author    = {Kachhadiya, Ridham and Patil, Dhanishtha and Anastasiu, David C.},
  title     = {Multi-Agent Cooperation for Traffic Safety Description and Analysis},
  booktitle = {Proceedings of the IEEE/CVF International Conference on Computer Vision Workshops},
  pages     = {5486--5494},
  year      = {2025}
}

@article{spatialagent2025,
  author  = {Huang, Hsiang-Wei and Cheng, Jen-Hao and Chen, Kuang-Ming and Yang, Cheng-Yen and Alattar, Bahaa and Lin, Yi-Ru and Kim, Pyongkun and Kim, Sangwon and Kim, Kwangju and Huang, Chung-I and Hwang, Jenq-Neng},
  title   = {Warehouse Spatial Question Answering with {LLM} Agent},
  journal = {arXiv preprint arXiv:2507.10778},
  year    = {2025}
}

@inproceedings{fisheye8k2023,
  author    = {Gochoo, Munkhjargal and Otgonbold, Munkh-Erdene and Ganbold, Erkhembayar and Hsieh, Jun-Wei and Chang, Ming-Ching and Chen, Ping-Yang and Dorj, Byambaa and Al Jassmi, Hamad and Batnasan, Ganzorig and Alnajjar, Fady and Abduljabbar, Mohammed and Lin, Fang-Pang},
  title     = {{FishEye8K}: A Benchmark and Dataset for Fisheye Camera Object Detection},
  booktitle = {Proceedings of the IEEE/CVF Conference on Computer Vision and Pattern Recognition Workshops},
  pages     = {5304--5312},
  year      = {2023}
}

@inproceedings{jaad2017,
  title={Are they going to cross? a benchmark dataset and baseline for pedestrian crosswalk behavior},
  author={Rasouli, Amir and Kotseruba, Iuliia and Tsotsos, John K},
  booktitle={2017 IEEE International Conference on Computer Vision Workshops (ICCVW)},
  pages={206--213},
  year={2017},
  organization={IEEE}
}

@inproceedings{pie2019,
  author    = {Rasouli, Amir and Kotseruba, Iuliia and Kunic, Toni and Tsotsos, John K.},
  title     = {{PIE}: A Large-Scale Dataset and Models for Pedestrian Intention Estimation and Trajectory Prediction},
  booktitle = {Proceedings of the IEEE/CVF International Conference on Computer Vision},
  pages     = {6261--6270},
  year      = {2019}
}

@article{vadr1,
  title={Vad-r1: Towards video anomaly reasoning via perception-to-cognition chain-of-thought},
  author={Huang, Chao and Wang, Benfeng and Wang, Wei and Wen, Jie and Liu, Chengliang and Shen, Li and Cao, Xiaochun},
  journal={Advances in neural information processing systems},
  volume={38},
  pages={118486--118518},
  year={2026}
}

@article{tadlocalization,
  author  = {Lv, Hui and Zhou, Chuanwei and Cui, Zhen and Xu, Chunyan and Li, Yong and Yang, Jian},
  title   = {Localizing Anomalies from Weakly-Labeled Videos},
  journal = {IEEE Transactions on Image Processing},
  volume  = {30},
  pages   = {4505--4515},
  year    = {2021},
  doi     = {10.1109/TIP.2021.3072863}
}

@article{accidentbench,
  title={Accidentbench: Benchmarking multimodal understanding and reasoning in vehicle accidents and beyond},
  author={Gu, Shangding and Wang, Xiaohan and Ying, Donghao and Zhao, Haoyu and Yang, Runing and Jin, Ming and Li, Boyi and Pavone, Marco and Yeung-Levy, Serena and Wang, Jun and others},
  journal={arXiv preprint arXiv:2509.26636},
  year={2025}
}

@article{tadbenchmark,
  title={TAD: A large-scale benchmark for traffic accidents detection from video surveillance},
  author={Xu, Yajun and Hu, Huan and Huang, Chuwen and Nan, Yibing and Liu, Yuyao and Wang, Kai and Liu, Zhaoxiang and Lian, Shiguo},
  journal={IEEE Access},
  volume={13},
  pages={2018--2033},
  year={2024},
  publisher={IEEE}
}

@article{sotad,
  author  = {Chen, Xingyuan and Xu, Huahu and Ruan, Mingyang and Bian, Minjie and Chen, Qishen and Huang, Yuzhe},
  title   = {{SO-TAD}: A Surveillance-Oriented Benchmark for Traffic Accident Detection},
  journal = {Neurocomputing},
  volume  = {618},
  pages   = {129061},
  year    = {2025},
  doi     = {10.1016/j.neucom.2024.129061}
}

@inproceedings{ucfcrime,
  author    = {Sultani, Waqas and Chen, Chen and Shah, Mubarak},
  title     = {Real-World Anomaly Detection in Surveillance Videos},
  booktitle = {Proceedings of the IEEE Conference on Computer Vision and Pattern Recognition},
  year      = {2018}
}

@misc{fetv2026,
  author       = {Abduljawad, Ahmed and Shaik, Nadeem Shahid and S S, Mohanrasu and Chang, Ming-Ching and Hsieh, Jun-Wei and Gochoo, Munkhjargal},
  title        = {{FETV}: Fisheye Traffic Event and Violation Dataset},
  year         = {2026},
  howpublished = {\url{https://github.com/MoyoG/FETV}}
}

@misc{psi2026,
  author       = {{ISE-ICE Lab}},
  title        = {{PSI-VQA}: {AI City Challenge 2026 Track 3} Out-of-Domain Pedestrian Intent VQA Dataset},
  year         = {2026},
  howpublished = {\url{https://huggingface.co/datasets/ise-ice-lab/PSI_VQA}}
}

@article{bertscore,
  title={Bertscore: Evaluating text generation with bert},
  author={Zhang, Tianyi and Kishore, Varsha and Wu, Felix and Weinberger, Kilian Q and Artzi, Yoav},
  journal={arXiv preprint arXiv:1904.09675},
  year={2019}
}

@article{yang2023revisiting,
  title={Revisiting auc-oriented adversarial training with loss-agnostic perturbations},
  author={Yang, Zhiyong and Xu, Qianqian and Hou, Wenzheng and Bao, Shilong and He, Yuan and Cao, Xiaochun and Huang, Qingming},
  journal={IEEE Transactions on Pattern Analysis and Machine Intelligence},
  volume={45},
  number={12},
  pages={15494--15511},
  year={2023},
  publisher={IEEE}
}

@article{hua2025openworldauc,
  title={Openworldauc: Towards unified evaluation and optimization for open-world prompt tuning},
  author={Hua, Cong and Xu, Qianqian and Yang, Zhiyong and Wang, Zitai and Bao, Shilong and Huang, Qingming},
  journal={arXiv preprint arXiv:2505.05180},
  year={2025}
}

@article{hua2024reconboost,
  title={Reconboost: Boosting can achieve modality reconcilement},
  author={Hua, Cong and Xu, Qianqian and Bao, Shilong and Yang, Zhiyong and Huang, Qingming},
  journal={arXiv preprint arXiv:2405.09321},
  year={2024}
}

@article{bao2025towards,
  title={Towards size-invariant salient object detection: A generic evaluation and optimization approach},
  author={Bao, Shilong and Xu, Qianqian and Li, Feiran and Han, Boyu and Yang, Zhiyong and Cao, Xiaochun and Huang, Qingming},
  journal={IEEE Transactions on Pattern Analysis and Machine Intelligence},
  year={2025},
  publisher={IEEE}
}

@article{bao2025aucpro,
  title={Aucpro: Auc-oriented provable robustness learning},
  author={Bao, Shilong and Xu, Qianqian and Yang, Zhiyong and He, Yuan and Cao, Xiaochun and Huang, Qingming},
  journal={IEEE Transactions on Pattern Analysis and Machine Intelligence},
  volume={47},
  number={6},
  pages={4579--4596},
  year={2025},
  publisher={IEEE}
}

@article{bao2024improved,
  title={Improved diversity-promoting collaborative metric learning for recommendation},
  author={Bao, Shilong and Xu, Qianqian and Yang, Zhiyong and He, Yuan and Cao, Xiaochun and Huang, Qingming},
  journal={IEEE Transactions on Pattern Analysis and Machine Intelligence},
  volume={46},
  number={12},
  pages={9004--9022},
  year={2024},
  publisher={IEEE}
}

@article{bao2022rethinking,
  title={Rethinking collaborative metric learning: Toward an efficient alternative without negative sampling},
  author={Bao, Shilong and Xu, Qianqian and Yang, Zhiyong and Cao, Xiaochun and Huang, Qingming},
  journal={IEEE Transactions on Pattern Analysis and Machine Intelligence},
  volume={45},
  number={1},
  pages={1017--1035},
  year={2022},
  publisher={IEEE}
}
\end{document}